\documentclass{article}

\usepackage[preprint]{corl_2026} 

\usepackage{booktabs}
\usepackage{amsmath}
\usepackage{amssymb}
\usepackage{amsthm}
\usepackage{subcaption}
\usepackage{xspace}
\usepackage{multirow}
\usepackage{graphicx}
\usepackage{wrapfig}
\usepackage{float}
\usepackage[most]{tcolorbox}
\usepackage{enumitem}

\newtcolorbox{promptbox}[1]{
  colback=gray!3,
  colframe=gray!45,
  title=#1,
  fonttitle=\bfseries,
  boxrule=0.5pt,
  arc=2pt,
  left=4pt,
  right=4pt,
  top=4pt,
  bottom=4pt
}

\graphicspath{{images/}}

\newcommand{\E}{\mathbb{E}}
\newcommand{\R}{\mathbb{R}}

\newcommand{\eprove}{$\epsilon$-ProVe\xspace}
\newcommand{\fearl}{FEARL\xspace}
\newcommand{\modC}{Module~C\xspace}
\newcommand{\modS}{Module~S\xspace}

\theoremstyle{plain}
\newtheorem{proposition}{Proposition}

\title{Verifiable Foundation Models for Robot Safety}

\author{
  Davide Corsi\thanks{Equal contribution.} \quad
  Kyungmin Kim$^{*}$ \quad
  Roy Fox \\
  University of California, Irvine \\
  \texttt{\{dcorsi, kyungk7, royf\}@uci.edu}
}

\begin{document}
\maketitle

\begin{abstract}
Deploying foundation models for robot control raises a central challenge: the expressive power that enables rich, multimodal perception also makes these models opaque and difficult to analyze formally, rendering them intractable for existing verification tools. In this paper, we present \fearl (Foundation-Enabled Assured Robot Learning), a framework that addresses this tension through a modular architectural decomposition. \fearl separates the policy into a large Controller (C) responsible for high-dimensional perception and task reasoning, and a small Safety module (S) that receives low-dimensional observations from dedicated safety sensors together with a bounded context embedding from C and produces the final action. Since many robot safety requirements, such as collision avoidance and workspace boundary constraints, can be expressed over these safety sensor observations, formal verification can be applied to S rather than to the full foundation-model backbone. This makes formal analysis tractable with existing tools while preserving the Controller's expressive power for task reasoning. To show that the decomposed policy remains capable of solving diverse tasks, we evaluate \fearl on three simulated robotic domains using multiple Controller backbones and training procedures, including pretrained off-the-shelf vision-language-action models. We further transfer the learned policy from one of our simulated tasks to a physical robot, suggesting that the low-dimensional safety interface supports practical sim-to-real transfer.
\end{abstract}


\section{Introduction}
\label{sec:intro}
Deploying robots in the real world ideally requires two capabilities that have historically been in tension: the ability to interpret rich, open-ended inputs such as natural language instructions and visual observations, and the ability to provide formal guarantees that the robot will not behave unsafely. Foundation models, including large language models (LLMs)~\cite{devlin2019bert,comanici2025gemini} and vision transformers~(ViTs)~\cite{dosovitskiy2020image}, have dramatically advanced the former, enabling robots to follow complex instructions and generalize across diverse environments, and have recently been integrated into end-to-end robot control pipelines as vision-language-action models (VLAs)~\cite{shukor2025smolvla, black2024pi0}. Yet their scale and complexity make them opaque to formal verification tools, which remain effective only for small, low-dimensional networks and do not scale to the size of modern foundation-model policies~\cite{wu2024marabou}.

A common solution is to train a large policy to be empirically safe through safety rewards, constrained reinforcement learning, or adversarial data augmentation~\cite{StAcAb20, YaSiJa22, achiam2017constrained}. However, empirical safety is not a certificate: extremely risk-sensitive systems (“five nines”) require infeasible amounts of empirical evaluation, and even then failures may appear outside the evaluation distribution. Alternatively, runtime shields provide a stronger last line of defense by filtering unsafe actions during deployment, but come with two concrete costs~\cite{corsi2024Verification}. First, evaluating safety constraints at every decision step is expensive, and finding a safe action in continuous domains with multiple constraints can grow prohibitively slow for high control frequencies. 
\begin{wrapfigure}{r}{0.5\textwidth}
  \vspace{-0.2em}
    \centering
    \includegraphics[width=\linewidth]{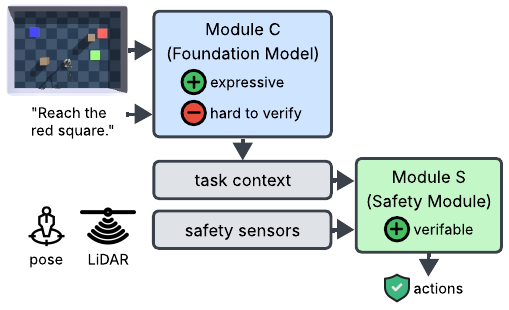}
    \vspace{-1.5em}
    \caption{Overview of the \fearl C/S decomposition: Controller (C) processes high-dimensional task inputs into low-dimensional context, while Safety module (S) uses this context and low-dimensional safety signals to produce the final, verifiable action.}
    \label{fig:intro_teaser}
   \vspace{-1.0em}
\end{wrapfigure}
Second, and more subtly, the replacement action returned by the shield gives no guarantee that it is the best alternative with respect to the original task objective, even if it is close to the original in some geometric sense~\cite{kim2025Realizable}.
Beyond these practical costs, shielding leaves a more fundamental question unanswered: can we determine, before deployment, which parts of the input domain the learned policy already handles safely, and which genuinely require runtime intervention? Without such a pre-deployment certificate, we cannot quantify how much the policy relies on the shield, nor bound the performance cost of its interventions. Verification is therefore the primary goal, with shielding reserved as a targeted fallback.

Motivated by this, we propose \fearl (Foundation-Enabled Assured Robot Learning), a framework that addresses this challenge through a principled architectural decomposition. The key observation is that many common robot safety requirements, such as collision avoidance and workspace boundary constraints, can be expressed over low-dimensional, semantically interpretable signals, such as LiDAR distances or robot pose, even when task-oriented behavior requires rich visual and language context. \fearl separates the agent into two modules: a large, expressive Controller (C) that processes high-dimensional inputs using pretrained foundation models, and a small Safety module (S) that receives the safety signals along with a low-dimensional context embedding from C and produces the final action (Figure~\ref{fig:intro_teaser}). Verification is applied only to S, making formal analysis tractable with existing tools while retaining the foundation model's role in perception and task reasoning.

Even with a compact, verifiable \modS, training a policy that satisfies all safety constraints across the entire input domain remains challenging in practice: corner cases and rarely visited regions of the safety sensor space may not receive sufficient training signal to guarantee safe behavior everywhere. In our pipeline, verification identifies regions of \modS’s input domain where the learned policy is guaranteed to satisfy the safety specification \cite{corsi2024Verification}. These certified regions are directly useful for deployment: actions in them can pass through unchanged, so shielding is reserved for inputs outside the certificate. Thus, verification certifies not only safety, but also where policy modification can occur. As formalized in our analysis, certified safe volume provides a coarse measure of shield-induced disruption: larger certified regions leave less room for shielding to alter behavior. While actual disruption also depends on the policy’s runtime visitation, certified volume gives an actionable development objective for low-disruption shielding: increase the certified safe region so that intervention is confined to a smaller uncertified region. When certified-region membership checks are cheaper than full shield queries, the same certificate can also reduce runtime overhead.

We evaluate \fearl on three robotic domains: a 2D playground, indoor navigation with a Hello Robot Stretch2, and outdoor navigation with a Unitree GO2, spanning different controller backbones, context representations, and training procedures. Across these domains, the two-module pipeline remains competitive, showing that routing final action selection through the compact safety module does not prevent effective task learning. 
For verification, we extend $\epsilon$-ProVe~\cite{marzari2024enumerating} to handle universally quantified context variables and action spaces, and certify large safe regions of the safety module’s input domain under multiple safety specifications. These large certified safe volumes are consistent with low-disruption shielding: verification-guided shielding achieves zero safety violations across all environments, while maintaining low override rates and small performance drops. We further transfer the indoor navigation policy to a physical Stretch robot without retraining.

We present \fearl with the following contributions: \textbf{(i)~A decomposition that makes foundation-model policies verifiable}, by separating the policy into a large expressive Controller (C) and a compact Safety module (S) that can be formally analyzed with existing tools; \textbf{(ii)~Verification-guided shielding with bounded policy disruption}, extending $\epsilon$-ProVe to handle universally quantified context variables and multiple safety constraints to certify large regions of \modS's input domain; \textbf{(iii)~Flexible controller backbones}, including an off-the-shelf VLA, showing the framework is compatible with diverse foundation-model pipelines; and \textbf{(iv)~Zero-shot sim-to-real transfer}, deploying the indoor navigation policy on a physical Stretch robot without retraining, demonstrating that both the task policy and the verified safety shield transfer directly from simulation to hardware.
\section{Related Work}
\label{sec:related}

\subsection{Foundation Models for Robot Control}
The use of large pretrained models for robot control has accelerated rapidly in recent years. Early work demonstrated that language models could be used as high-level planners, decomposing natural language instructions into sequences of executable primitives~\citep{ahn2022saycan, huang2022language}. Subsequent efforts moved toward end-to-end approaches, training vision-language-action (VLA) models that directly map image observations and language instructions to robot actions~\citep{brohan2023rt2, kim2024openvla}. More recent systems such as $\pi_0$~\citep{black2024pi0} combine large pretrained vision-language backbones with flow-matching action heads, achieving strong generalization across diverse manipulation tasks. Alongside these, imitation learning from human demonstrations has proven effective for grounding high-level semantic understanding into low-level motor control~\citep{chi2025diffusion, zhao2023learning}. Despite impressive empirical performance, none of these approaches provide formal guarantees on the behavior of the deployed policy. Their scale and architectural complexity place them far beyond the reach of existing neural network verification tools, leaving safety entirely to empirical evaluation.

\subsection{Verification of Neural Networks and Shielding}
Formal verification of deep neural networks (DNNs) asks whether a network satisfies a given safety specification for every possible input in a specified domain. More formally, given a DNN $\mathcal{N}$, a precondition $\mathcal{P}$ on its inputs, and a postcondition $\mathcal{Q}$ on its outputs, a verifier determines whether
\begin{equation}
    \exists\, x \;:\; \mathcal{P}(x) \;\wedge\; \mathcal{Q}(\mathcal{N}(x)),
    \label{eq:verification}
\end{equation}
holds~\citep{katz2017reluplex}. In practice, $\mathcal{P}$ encodes domain-specific knowledge, for instance restricting the input to a dangerous region of the state space, while $\mathcal{Q}$ encodes the negation of the desired behavior. A verifier that returns \textsc{unsat} for Eq.~\ref{eq:verification} therefore certifies that the network behaves correctly on every input in the domain of interest. This problem is NP-complete in general, and a number of dedicated solvers have been developed to tackle it, including Marabou~\citep{wu2024marabou}, $\alpha,\beta$-CROWN~\citep{zhang2018efficient}, and $\epsilon$-ProVe~\citep{marzari2024enumerating}, the latter of which enumerates provably safe regions of the input domain with probabilistic volume guarantees. Despite these advances, all existing verifiers remain tractable only for relatively small networks, far below the scale of any modern foundation model.

Runtime shielding offers an alternative approach to verification by intercepting unsafe actions at deployment time~\citep{alshiekh2018safe, Lazarus2020Runtime}, but introduces policy disruption and per-step computational overhead. Verification-guided shielding~\citep{corsi2024Verification} combines both techniques by certifying that the shield is unnecessary over most of the input domain and activating it only in the small uncertified region. \fearl builds on this pipeline, extending it to the foundation-model setting by decoupling a small, verifiable module from a large pretrained backbone.
\section{The FEARL Architecture}
\label{sec:architecture}
The central design goal of \fearl is to make formal verification compatible with foundation-model-empowered robot policies. Since end-to-end policies with foundation-model backbones entangle safety-critical decisions with large perception, language, or multimodal modules, verifying them is infeasible with current tools.
\fearl decouples these aspects of safe control.
\subsection{Decomposition Principle}
\label{sec:decomposition}

Consider a robot observing at time $t$ the triplet $o_t = (I_t,\, g,\, s_t)$, where $I_t$ is a visual observation, $g$ is a natural language task description, and $s_t \in \mathcal{S} \subset \R^d$ is a low-dimensional \emph{safety sensor} vector, such as LiDAR distances or robot pose, with well defined semantics.

The key insight underlying \fearl is that many robot safety constraints can be expressed entirely in terms of the robot's local physical relationship to its environment, which is captured by $s_t$, rather than the full semantic content of raw pixels $I_t$ or the language instruction $g$. For example, deciding which object to approach may require visual and language understanding, but deciding whether moving forward would violate a collision-avoidance constraint can be determined from local range or pose measurements. \fearl exploits this insight by explicitly separating semantic context from safety-critical action selection. \modC processes the high-dimensional inputs $(I_t, g)$ into a small context embedding, while \modS receives this embedding together with the safety sensor vector $s_t$ and produces the final action.

\subsection{Module C: Foundation-Model Backbone}
\label{sec:moduleC}

\modC is the controller module of \fearl. Its role is to process high-dimensional task inputs, such as images, language instructions, or other domain-specific observations, and produce a compact context embedding $z_t \in \mathcal{Z} \subset \mathbb{R}^{d_c}$ that is passed to \modS. We intentionally leave the internal architecture of \modC unconstrained: it may be a custom-trained encoder, a fine-tuned foundation model, an off-the-shelf vision-language-action model, or a domain-specific perception pipeline. The only interface requirement is that \modC outputs a bounded context vector $z_t \in \mathcal{Z} = [-1,1]^{d_c}$, enforced in practice by applying a tanh activation to the output of \modC. This boundedness is not merely a design preference: it is a formal requirement for verification, since \eprove enumerates safe regions over the full input domain $\mathcal{X} = \mathcal{S} \times \mathcal{Z}$ of \modS, which must be finite for the procedure to be well-defined. 

The semantics of $z_t$ does not need to be prescribed in advance: when \modC and \modS are trained jointly, \modC naturally tends to encode task-relevant information into $z_t$, while \modS learns to combine it with $s_t$ to produce safe and effective actions. The embedding may therefore be a learned latent representation obtained by projecting the hidden state of a foundation model through a small adapter, or a structured hand-designed descriptor such as robot and goal locations. In either case, \modC is not tied to a specific backbone, modality, or training procedure; different domains can instantiate it differently as long as they expose the same bounded context interface. In our experiments we instantiate \modC in several ways across domains (see able~\ref{tab:setup}), demonstrating that the same \fearl verification pipeline applies regardless of the particular backbone or training procedure.

\subsection{Module S: Small Verifiable Safety-Action Module}
\label{sec:moduleS}

\modS is the safety-action module of \fearl. It takes as input the concatenation 
$x_t = [s_t;\, z_t] \in \mathcal{X}$, where $s_t \in \mathcal{S} \subset \mathbb{R}^d$ is the safety sensor vector and $z_t \in \mathcal{Z} \subset \mathbb{R}^{d_c}$ is the bounded context embedding from \modC, and produces the action distribution from which the final action is selected. \modS is kept deliberately small so that it can be formally analyzed by existing neural network verification tools~\cite{wu2024marabou, SuKhSh19}. In our implementation, it is a two-layer MLP with 32 units per layer, which keeps verification tractable for \eprove's probabilistic enumeration of axis-aligned hyperrectangular safe regions~\cite{marzari2024enumerating}.

The safety sensor vector $s_t$ provides the interpretable interface on which the verification requirements are formalized. Its components are low-dimensional, human-readable, and physically meaningful, such as LiDAR distances, robot pose, or distances to workspace boundaries, allowing safety requirements to be stated directly in terms of the robot's measurable physical state. Formally, a safety property takes the form
\[
  s_t \in \mathcal{R}
  \;\Longrightarrow\;
  \pi_S(\cdot \mid x_t = (s_t, z_t)) \notin \mathcal{Y}_\text{unsafe}(\mathcal{R}),
\]
where $\mathcal{R} \subset \mathcal{S}$ is a bounded region of the safety sensor space and $\mathcal{Y}_\text{unsafe}(\mathcal{R})$ is the set of \modS outputs that are unsafe in that region. For discrete control, this may mean that the highest-probability action belongs to a prohibited set $\mathcal{A}_\text{unsafe} \subseteq \mathcal{A}$; for a continuous controller it may mean that the action or distribution parameters fall outside a safe output set. Critically, the property must hold for \emph{all} admissible context embeddings $z_t \in \mathcal{Z}$, ensuring that \modS never selects an unsafe action in the specified sensor region regardless of what the foundation model perceives or infers from the high-dimensional inputs, and of any hallucinations it may have. While our experiments use a discrete action set $\mathcal{A}$, the framework is not limited to discrete control.

\section{Enabling Verification via Decomposition}
\label{sec:verification}

Our safety assurance pipeline combines \emph{offline} neural network verification
with \emph{selective} runtime shielding, following the verification-guided shielding paradigm \cite{corsi2024Verification}. 

\paragraph{Offline phase.}  
We apply \eprove to \modS to partition the joint input domain
$\mathcal{X} = \mathcal{S} \times \mathcal{Z}$ into:
$(i)$~a \emph{provably safe region} $\mathcal{X}_\text{certified}$,
where \modS is formally certified to satisfy all safety properties; and
$(ii)$~an \emph{uncertified region} $\mathcal{X}_\text{uncertified} = \mathcal{X} \setminus \mathcal{X}_\text{certified}$,
where no formal guarantee can be established.
This analysis is computationally expensive but is performed \emph{once} before deployment,
so it does not affect runtime performance.

\paragraph{Runtime phase.}
A lightweight shield monitors the current state $s_t$. If $(s_t, z_t) \in \mathcal{X}_\text{certified}$, the shield is bypassed and \modS's action is executed directly. If $(s_t, z_t) \in \mathcal{X}_\text{uncertified}$ and the action turns out unsafe, the shield intervenes and replaces the proposed action with a certified safe alternative. This selective activation is the key to low override rates: because the majority of the state space is certified offline, the shield rarely needs to act at runtime.

\subsection{Theoretical Guarantees}
\label{sec:theory}

We say that a verifier with normalized input-space measure $\mu$ (typically standard Borel) certifies $\mathcal{X}_\text{certified}$ as $(\delta,\rho)$-safe if, with probability at least $\delta$ over the verifier's stochastic execution,
the safety false-positive rate $\frac{\mu(\mathcal{X}_\text{certified} \cap \mathcal{X}_\text{viol})}{\mu(\mathcal{X}_\text{certified})}$
is at most $\rho$, where $\mathcal{X}_\text{viol}$ is the set of inputs where \modS truly violates safety.
Let $\bar{\pi}$ denote the $\mathcal{X}_\text{uncertified}$-shielded policy,
and $d_T^{\pi}(s) = \frac{1}{T}\sum_{t=0}^{T-1}f^\pi_t(s)$
be the finite-horizon state occupancy density.
Write $\bar{d}_T^\pi = \sup_s d_T^\pi(s)$ for the maximum occupancy density.

The following three propositions, proved in the appendices, place bounds on the shielded policy's safety violation rate, shield activation rate, and value deterioration compared to the unshielded policy.

\begin{proposition}[Probabilistic Safety Guarantee]
\label{prop:safety}
With probability at least $\delta$, the safety violation rate of the shielded policy satisfies
\[
  \beta_T(\bar{\pi})
  \;\leq\;
  \rho\,\bar{d}_T^{\,\bar{\pi}}\,\mu(\mathcal{X}_\text{certified}).
\]
\end{proposition}

\begin{proposition}[Shield Activation Bound]
\label{prop:activation}
The expected shield activation rate  $\alpha_T(\pi) = d_T^\pi(\mathcal{X}_\text{uncertified})$ satisfies
\[
  \alpha_T(\pi) \;\leq\; \bar{d}_T^\pi\,\mu(\mathcal{X}_\text{uncertified}),
\]
\end{proposition}

\begin{proposition}[Value Deterioration Bound]
\label{prop:value}
Let $R_{\max}$ be the maximum per-episode return.
The shielded policy's $T$-step value satisfies
\[
  J_T^{\bar{\pi}} \;\geq\; J_T^\pi \;-\; T\,\bar{d}_T^\pi\,\mu(\mathcal{X}_\text{uncertified})\,R_{\max}.
\]
\end{proposition}

All three results follow from the definition of the occupancy measure and the Performance Difference Lemma~\citep{achiam2017constrained}. Together, Propositions~\ref{prop:activation} and~\ref{prop:value} establish
a direct link between the \emph{volume of the uncertified region}
and both the shield's runtime impact and the performance cost of shielding.
Minimizing $\mu(\mathcal{X}_\text{uncertified})$ through better training of \modS is
therefore simultaneously a safety and a performance objective.

\section{Experiments}
\label{sec:experiments}

We evaluate \fearl across domains that vary in task complexity, sensing modality, robot form factor, and \modC instantiation. The experiments are organized around two central objectives. First, we assess whether the C/S decomposition supports effective policy learning: despite delegating all safety-critical decisions to the small \modS, the agent should still achieve competitive task performance across diverse environments and training procedures. Second, we assess whether \modS can be formally verified by existing formal methods to certify large portions of its input domain as provably safe and enabling assured runtime safety through verification-guided shielding. Together, these two objectives test the core claim of \fearl: that expressive foundation-model perception and formal safety guarantees are not in conflict, but can coexist within a single deployable agent.

\subsection{Experimental Setup}
\label{sec:setup}

\begin{figure}[t]
\centering

\begin{minipage}[t]{0.59\linewidth}
\vspace{0pt}
\centering
\captionof{table}{%
    Environment, architecture, and training configuration.
    $d_s$: safety sensor dimension;
    $d_c$: context embedding dimension;
    $|\mathcal{A}|$: action space size.
}
\label{tab:setup}
\scriptsize
\setlength{\tabcolsep}{2.5pt}
\resizebox{\linewidth}{!}{%
\begin{tabular}{lccclll}
\toprule
Environment & $d_s$ & $d_c$ & $|\mathcal{A}|$ & Embedding & Module C & Training \\
\midrule
Playground           & 2  & 64  & 4 & Learned & BERT+ViT & PPO+LoRA \\
Indoor Nav           & 11 & 12  & 3 & Structured & LLM+ViT  & SFT$\to$PPO \\
Indoor Nav (VLA)          & 11 & 16 & 3 & Learned & SmolVLA  & PPO+LoRA \\
Outdoor Nav          & 3  & 128 & 4 & Learned & ViT      & DAgger \\
\bottomrule
\end{tabular}%
}
\end{minipage}
\hfill
\begin{minipage}[t]{0.4\linewidth}
\vspace{0pt}
\centering
\setcounter{subfigure}{0}
\captionsetup[subfigure]{justification=centering,singlelinecheck=false,font=scriptsize}

\begin{subfigure}[t]{0.32\linewidth}
    \centering
    \includegraphics[width=\linewidth]{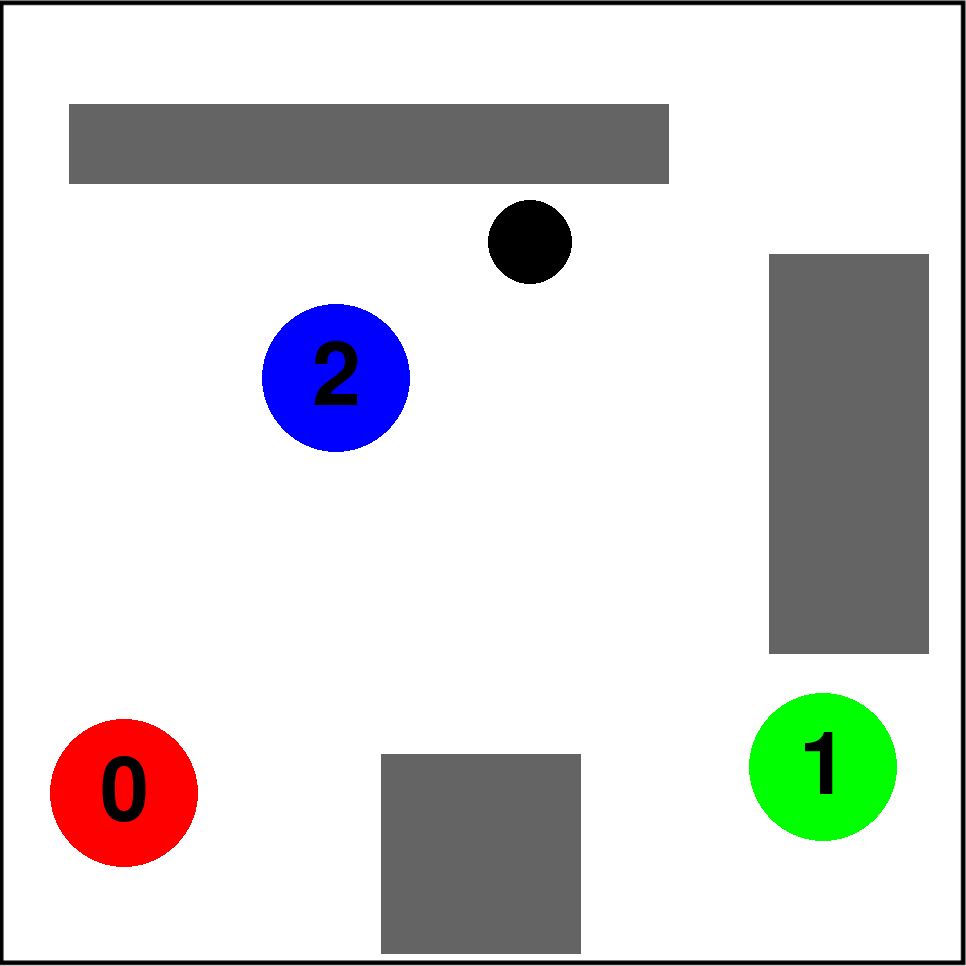}
    \caption{Playground}
\end{subfigure}
\hfill
\begin{subfigure}[t]{0.32\linewidth}
    \centering
    \includegraphics[width=\linewidth]{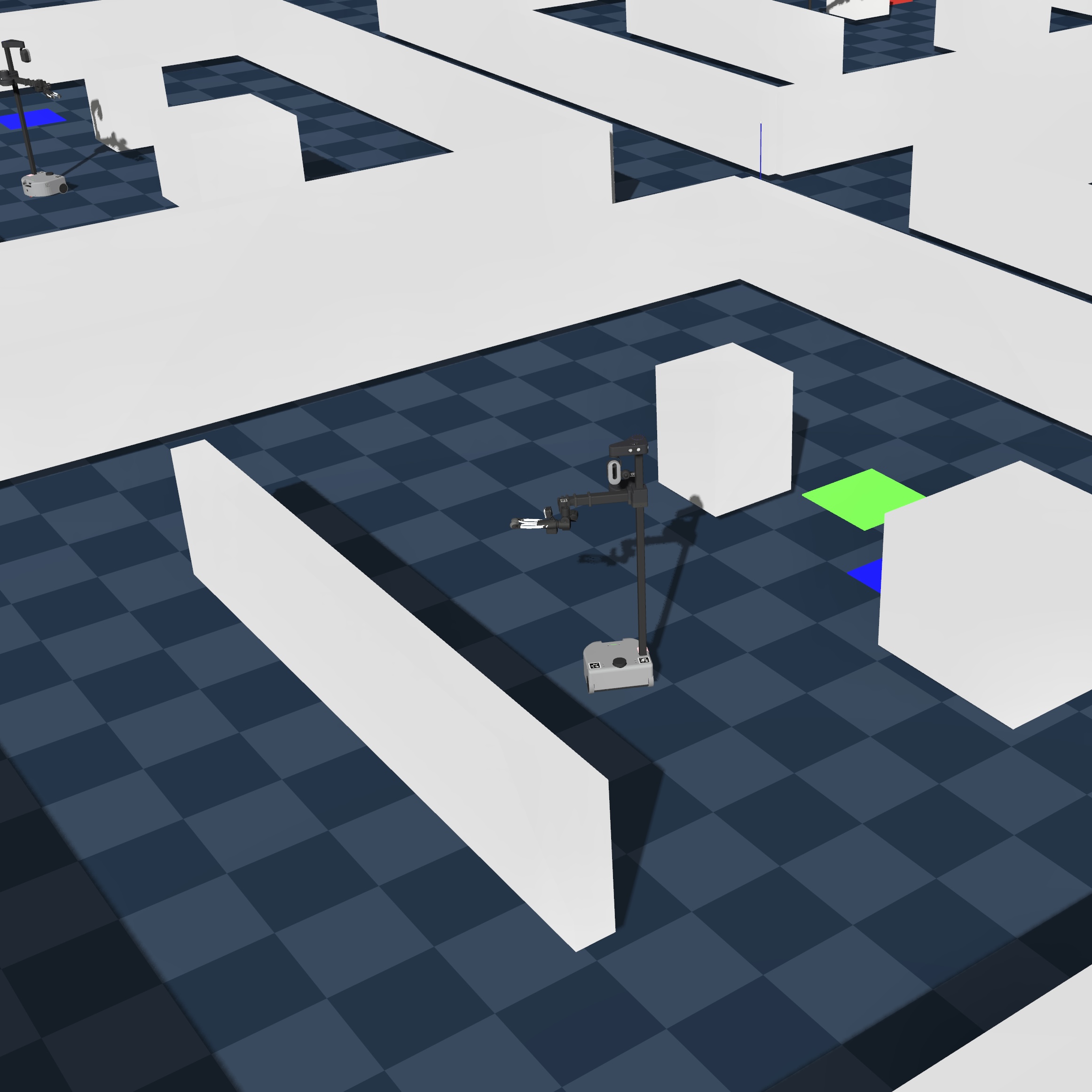}
    \caption{Indoor \\ (Stretch 2)}
\end{subfigure}
\hfill
\begin{subfigure}[t]{0.32\linewidth}
    \centering
    \includegraphics[width=\linewidth]{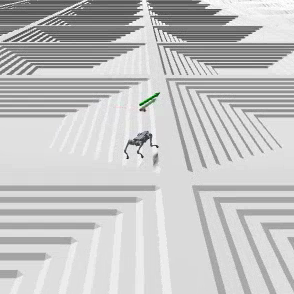}
    \caption{Outdoor \\ (Unitree GO2)}
\end{subfigure}

\captionof{figure}{Three environments across different robot platforms.}
\label{fig:environments}
\end{minipage}
\vspace{-5mm}
\end{figure}

\noindent\textbf{Environments.}
We evaluate \fearl on the domains summarized in Table~\ref{tab:setup} and shown in 
Figure~\ref{fig:environments}. \textbf{(1)~Playground} is a 2D synthetic task in a bounded arena that serves as an integration test for the full pipeline; its safety observation is the agent's normalized position $s_t\in[0,1]^2$. \textbf{(2)~Indoor Navigation} is evaluated on a Hello Robot Stretch~2~\citep{kemp2022design} navigating toward language-specified goal regions while avoiding obstacles. The policy observes the full arena image together with a language instruction, and its safety observation 
is an 11-ray LiDAR scan $s_t\in[0,3]^{11}$ over a $180^\circ$ forward field of view.
\textbf{(3)~Outdoor Navigation} is a target-reaching task evaluated on a Unitree GO2 
quadruped~\citep{rudin2022learning}; its safety observation is the robot's normalized planar pose $s_t=[s_x,s_y,s_\psi]\in[0,1]^3$. Full environment details are provided in the appendix.

\noindent\textbf{Architecture and training procedure.}
Table~\ref{tab:setup} summarizes the \modC instantiations and training procedures used in our experiments. 
We vary the controller architecture, context representation, and training procedure across domains. The controller backbones include BERT~\cite{devlin2019bert}+ViT~\cite{dosovitskiy2020image}, a custom LLM+ViT, a ViT-only model, and SmolVLA~\cite{shukor2025smolvla}, an off-the-shelf VLA-style architecture.
Depending on the domain, we train the \fearl policy with PPO~\cite{schulman2017proximal}+LoRA~\cite{hu2022lora}, supervised fine-tuning (SFT) followed by PPO, or DAgger~\cite{ross2011reduction}. 
The controller provides either free-form bounded embeddings or structured task descriptors as the context input $z_t$ to \modS.
For SmolVLA, we map the representation immediately before the VLA action head through a small MLP adapter to produce $z_t$.
Together, these instantiations illustrate that \fearl can support different controller designs, including standard VLA-style backbones, as long as they expose a bounded context representation to the safety-action module. Additional implementation details are provided in the appendix.

\noindent\textbf{Safety specifications.}
For each environment, safety requirements are expressed as safety-sensor-conditioned constraints on \modS: boundary avoidance for the Playground; LiDAR-based obstacle avoidance (5 properties) for Indoor Navigation; and boundary- and orientation-aware motion constraints (10 properties) for Outdoor Navigation. Full specifications can be found in the appendix.

\noindent\textbf{Evaluation.}
All policies are tested over 100 evaluation episodes over different seeds. We report success rate and collision/violation rate before and after shielding. For verification, we use an extended version of \eprove that supports universally quantified context variables and multiple simultaneous safety constraints on discrete action spaces, with a point-cloud size of 3500 and a depth limit of 20, targeting at most $\rho = 0.5\%$ false positives at confidence $\delta = 99.99\%$, unless otherwise specified.

\subsection{Task Performance of the Decomposed Policy} \label{sec:exp_performance}
Table~\ref{tab:shielding} reports the unshielded performance of the \fearl policies after training. Playground achieves perfect success ($100\%$) with no violations, indicating that the C/S decomposition does not impede learning in a simple navigation setting. In Indoor Navigation, the custom \modC policy achieves $73.1\%$ success with a violation rate of $1.3\%$, reflecting the challenge of safely controlling the Stretch platform given higher-dimensional LiDAR-based safety observations. The SmolVLA-based variant achieves more than 80\% success with a violation rate of less than 3\%, suggesting that the same \fearl interface can support a standard VLA-style controller in addition to the custom controller. 
We also observe qualitatively that the SmolVLA-based policy can follow out-of-distribution paraphrased instructions, suggesting that the bounded context interface can preserve useful language information from the VLA backbone. Outdoor Navigation achieves $89.6\%$ success with a violation rate of $1.6\%$.

These results show that the decomposed architecture can learn effective task policies before any runtime shielding is applied. This is important because \fearl's verification mechanism would not be useful if the architectural decomposition itself prevented the policy from solving the task. 

\subsection{Does the Decomposition Enable Verification?}
\label{sec:exp_verification}

Table~\ref{tab:verification} summarizes offline verification results from applying \eprove to \modS. Importantly, verification is performed only on the small safety-action module, not on the full foundation-model backbone. Thus, these results directly test whether the C/S decomposition yields a tractable target while certifying a substantial proportion of the input domain. For Playground ($d_s=2$), \eprove certifies $99.4\%$ of the input domain as provably safe in $22.7$ seconds. For the Indoor Navigation ($d_s=11$), the 11-dimensional LiDAR safety space leads to longer verification time ($1521.6$\,s) and a smaller certified fraction ($78.9\%$), though most of the domain is still certified. For Outdoor Navigation ($d_s=3$), \eprove certifies $82.7\%$ of the domain in $3271.2$\,s.  We highlight that verification may take minutes to hours depending on the domain and specification, but because this cost is incurred offline before deployment, it does not slow down the robot’s control loop.

\subsection{Can Verification Guide Low-Disruption Shielding?}
\label{sec:exp_shielding}

Table~\ref{tab:shielding} reports shielded performance and shield statistics. This experiment evaluates whether the certified safe region obtained from verifying \modS provides a meaningful bound on the behavioral impact of runtime shielding. In verification-guided shielding, actions in certified states are guaranteed safe and therefore passed through unchanged; shielding is only needed in the uncertified region. Thus, the certified safe volume is useful as a guarantee that policy modification is confined to a bounded portion of the input domain.

Across all environments, verification-guided shielding achieves \emph{zero safety violations} with minimal shield overrides and policy disruption, suggesting that the certified region covers the vast majority of states visited by the learned policy at deployment time.
These results are consistent with the large certified safe volumes in Table~\ref{tab:verification} and with Proposition~\ref{prop:value}: most of the domain is certified safe, leaving only a limited region where the shield can alter the learned policy.

The certified volume alone does not fully determine the override rate. As Proposition~\ref{prop:activation} suggests, shield usage also depends on the policy’s occupancy of the uncertified region and on how often unsafe actions are proposed there. This explains why Indoor Navigation can have a smaller certified volume than Outdoor Navigation while still exhibiting a lower override rate.

\begin{table}[t]
\centering
\small
\begin{minipage}[t]{0.52\linewidth}
\centering
\captionof{table}{Performance with and without shielding, averaged over three seeds. ``SR''/``VR'' denote success/violation rates (\%). ``Override rate'' is the fraction of steps where the proposed action is replaced.}
\label{tab:shielding}
\setlength{\tabcolsep}{4.5pt}
\resizebox{\linewidth}{!}{%
\begin{tabular}{lcc cc c}
\toprule
& \multicolumn{2}{c}{Unshielded} & \multicolumn{2}{c}{Shielded} &  \multirow{2}{*}{\shortstack{Override\\rate (\%)}} \\
\cmidrule(lr){2-3}\cmidrule(lr){4-5}
Environment & SR & VR & SR & VR & \\
\midrule
Playground                  & 100.0 & 0.0 & 100.0 & \textbf{0.0} & 0.09 \\
Indoor Nav-Custom           & 73.1 & 1.3 &  71.3 & \textbf{0.0} & 0.12 \\
Indoor Nav-VLA              & 80.1  & 2.7 & 68.2 & \textbf{0.0} & 1.13          \\
Outdoor Nav                 &  89.6 & 1.6 &  87.2 & \textbf{0.0} & 0.74 \\
\bottomrule
\end{tabular}%
}
\end{minipage}
\hfill
\begin{minipage}[t]{0.45\linewidth}
\centering
\caption{
Offline verification using \eprove. ``Certified volume'' is the certified proportion of the input domain; time is total offline computation.
}
\label{tab:verification}
\setlength{\tabcolsep}{3.5pt}
\resizebox{\linewidth}{!}{
\begin{tabular}{lcc}
\toprule
Environment & Certified volume & Time (s) \\
\midrule
Playground            & 99.4\% &   22.7  \\
Indoor Nav-Custom  & 78.9\% & 1521.6  \\
Indoor Nav-VLA  & 91.2\%   & 2690.8 \\
Outdoor Nav & 82.7\% & 3271.2  \\
\bottomrule
\end{tabular}
}
\end{minipage}
\end{table}

\subsection{Sim-to-Real Transfer}
\label{sec:sim2real}

\begin{wrapfigure}{r}{0.35\textwidth}
    \centering
    \includegraphics[width=\linewidth]{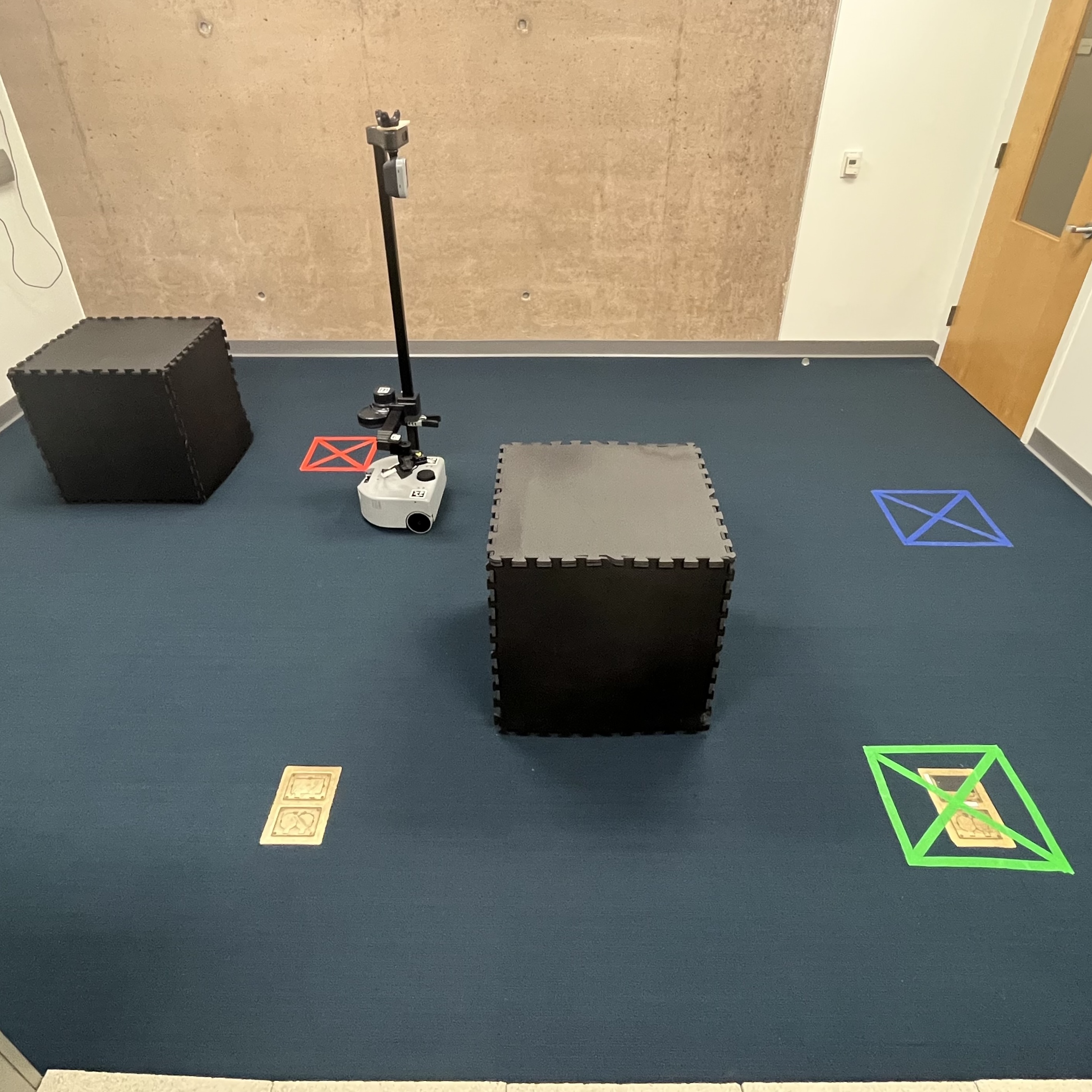}
    \caption{Real-world indoor navigation setup.}
    \label{fig:real_arena}
   \vspace{-1.0em}
\end{wrapfigure}

We further validate \fearl's sim-to-real transfer by deploying a physical Hello Robot Stretch~2~\cite{kemp2022design} in a real indoor environment, shown in Figure~\ref{fig:real_arena}. For this experiment, we adopt a mapless navigation variant of the Indoor Navigation task, where the agent has no access to a bird's-eye view of the arena and must navigate using only a language prompt processed by a foundation model and a local LiDAR-based safety observation, following an architecture similar to what is shown in the previous sections, but adapted to this reduced input space. 
Offline verification of \modS certifies $77.1\%$ of the input domain as provably safe, with hyperparameters adjusted from the simulation experiments for tractability ($n{=}100$ points, $\rho{=}0.5\%$, $\delta{=}97.6\%$) while remaining comparable.
In real-world rollouts over 18 episodes, the shielded policy achieves a $61.1\%$ success rate with zero collisions, while disabling the shield results in the same success rate but $27.8\%$ collision rate. The lower success rate compared to simulation is consistent with expected sim-to-real distribution shift in the task-level inputs, such as localization drift and sensor noise. Crucially, the safety mechanism remains fully effective: the shield activates in $7.8\%$ of steps, consistent with the $22.9\%$ uncertified volume identified offline, confirming that the verified safety interface transfers to hardware without retraining or re-verification. These results demonstrate that the C/S decomposition can be transfered to the real robot and, although performance may degrade under distribution shift, the safety guarantees are preserved by the verified shield, whose input remains within the requirements bounds as long as the safety sensors are properly calibrated.

\section{Discussion and Conclusion}
\label{sec:conclusion}

We have presented \fearl, an architecture for building foundation-model-based robot controllers with formal safety certificates. The C/S decomposition brings together two goals that are often difficult to reconcile: the expressive behavior of a large pretrained model and safety certificates obtainable with existing verification tools. 
Across three simulated environments, verification-guided selective shielding achieves zero runtime safety violations while keeping shield overrides and policy disruption low.
The formal bound on value deterioration (Proposition~\ref{prop:value}) explains why: shielding can degrade task performance only through interventions outside the certified region, which remains small for our trained \modS.

\noindent\textbf{Limitations and future work.}
Our current safety specifications focus on constraints observable through low-dimensional safety sensors, such as LiDAR or robot pose. This excludes richer semantic safety requirements that must be inferred from high-dimensional observations, such as recognizing that a glass cup should not be placed near the edge of a table. Extending \fearl to such semantic safety constraints, as well as richer temporal or multi-agent specifications, is an important direction for future work. More broadly, \fearl suggests a path toward verification-guided policy improvement: verification can identify when a learned policy is globally safe under a given specification and can generate safety counterexamples for adversarial training. As robot foundation models continue to improve, \fearl provides a way to combine expressive task reasoning with a stable, formally analyzable safety interface.

\clearpage
\clearpage 
\bibliography{fearl}
\clearpage \appendix

\section{Proof of Propositions}
\label{app:proofs}

\paragraph{Proof of Proposition~1 (Probabilistic Safety Guarantee).}
By definition,
$\beta_T(\bar{\pi})
 = \int_{\mathcal{X}_\text{certified} \cap \mathcal{X}_\text{viol}}
   d_T^{\bar{\pi}}\, ds$.
Bounding the integrand by the supremum gives
$\beta_T(\bar{\pi})
 \leq \bar{d}_T^{\bar{\pi}}\,
       \mu(\mathcal{X}_\text{certified}\cap\mathcal{X}_\text{viol})$.
The $(\delta,\rho)$-safety guarantee yields
$\mu(\mathcal{X}_\text{certified}\cap\mathcal{X}_\text{viol})
 \leq \rho\,\mu(\mathcal{X}_\text{certified})$
with probability at least $\delta$, giving the stated bound. \hfill$\square$

\paragraph{Proof of Proposition~2 (Shield Activation Bound).}
$\alpha_T(\pi)
 = \int_{\mathcal{X}_\text{uncertified}} d_T^\pi\, ds
 \leq \bar{d}_T^\pi\,\mu(\mathcal{X}_\text{uncertified})$.
\hfill$\square$

\paragraph{Proof of Proposition~3 (Value Deterioration Bound).}
By the Performance Difference Lemma~\citep{achiam2017constrained},
$J_T^\pi - J_T^{\bar{\pi}}
 = \E_{\xi\sim p_\pi}\!\bigl[\sum_{t=0}^{T-1} A_t^{\bar{\pi}}(s_t,a_t)\bigr]$.
The shield modifies actions only when $s_t\in\mathcal{X}_\text{uncertified}$,
so $A_t^{\bar{\pi}}(s_t,a_t)=0$ whenever $s_t\in\mathcal{X}_\text{certified}$.
Bounding the advantage by $R_\text{max}$ and applying Proposition~2
gives $J_T^\pi - J_T^{\bar{\pi}}
 \leq T\,\bar{d}_T^\pi\,\mu(\mathcal{X}_\text{uncertified})\,R_\text{max}$,
which rearranges to the stated lower bound on $J_T^{\bar{\pi}}$. \hfill$\square$

\section{Environment Details}
\label{app:env_details}

This section provides complete specifications for each environment used in our
experiments, including observation and action spaces, dynamics, reward functions,
and episode termination conditions.

\subsection{Playground}
\label{app:env_playground}

\noindent\textbf{Setup.}
Playground is a 2D synthetic task in a $1048\times1048$\,px bounded arena.
The agent is a circular body of radius $40$\,px ($\approx 0.038$ in normalized
coordinates).
At each step, one of four directional actions displaces the agent by $10$\,px;
both position and displacement are divided by $1048$ to obtain the normalized
representation.

\noindent\textbf{Observations.}
At each step $t$ the agent receives
$o_t = (I_t, g, s_t)$,
where $I_t \in \R^{96\times96\times3}$ is a top-down RGB image,
$g$ is a natural-language task description identifying one of three colored
goal regions, and the safety observation is the agent's normalized 2D position:
\[
  s_t = (s_x, s_y) \in [0,1]^2.
\]

\noindent\textbf{Action space.}
$\mathcal{A} = \{\textsc{Right}(0),\,\textsc{Left}(1),\,\textsc{Up}(2),\,\textsc{Down}(3)\}$.
Each action induces a fixed $10$\,px displacement in the corresponding cardinal
direction, producing a maximum step size of $10/1048 \approx 0.010$ in
normalized coordinates.

\noindent\textbf{Dynamics.}
$s_{t+1} = \mathrm{clip}(s_t + \delta \cdot a_t,\,[0,1]^2)$,
where $\delta = 10/1048$.
If the unclipped position would exit the arena, the episode terminates with an
out-of-bounds penalty.

\noindent\textbf{Reward and termination.}
The agent receives $+1.0$ for reaching the correct target, $-1.0$ for going
out of bounds, $-0.2$ for hitting an obstacle, and a step penalty of $-0.01$
to encourage efficiency.
Episodes terminate upon goal completion, a boundary or obstacle violation,
or after $T{=}500$ steps.

\subsection{Indoor Navigation}
\label{app:env_indoor}

\noindent\textbf{Setup.}
Indoor Navigation is evaluated on a simulated Hello Robot Stretch~2~\citep{kemp2022design}
in a $6\times6$\,m workspace (coordinates $[-3,3]\times[-3,3]$\,m) bounded by
walls of thickness $0.15$\,m.
The simulation uses a physics time step of $0.05$\,s with an action repeat of
$5$, giving $0.25$\,s of simulated time per environment step.
The robot follows differential-drive dynamics with wheel radius $0.05$\,m and
maximum wheel velocity $5$\,rad/s, yielding a maximum forward displacement of
approximately $0.063$\,m per environment step.

\noindent\textbf{Observations.}
The agent receives a top-down RGB image $I_t \in \R^{256\times256\times3}$ of
the full arena together with a language instruction $g$ specifying one of three
colored goal regions.
The safety observation is an 11-ray LiDAR scan over a $180^\circ$ forward field
of view:
\[
  s_t \in [0,3]^{11},
\]
where each entry gives the range in meters to the nearest obstacle along the
corresponding ray; values near $0$ indicate imminent collision.

\noindent\textbf{Action space.}
$\mathcal{A} = \{\textsc{Turn-Left}(0),\,\textsc{Turn-Right}(1),\,\textsc{Forward}(2)\}$.
Each action is executed as a pair of wheel velocities: Turn-Left ($v_l=-5.0$,
$v_r=5.0$), Turn-Right ($v_l=5.0$, $v_r=-5.0$), and Forward ($v_l=v_r=5.0$)
in units of rad/s.

\noindent\textbf{Dynamics.}
The robot follows standard unicycle kinematics:
$\dot{x} = \frac{r}{2}(v_r+v_l)\cos\theta$,
$\dot{y} = \frac{r}{2}(v_r+v_l)\sin\theta$,
$\dot{\theta} = \frac{r}{L}(v_r-v_l)$,
where $r=0.05$\,m is the wheel radius and $L$ is the wheelbase.

\noindent\textbf{Module C variants.}
This environment is evaluated with two distinct \modC instantiations
(see Section~\ref{app:arch_training} for training details).
\begin{enumerate}
  \item \textbf{Custom \modC (MDP):} An LLM and a ViT are trained by
    supervised fine-tuning to produce a structured $12$-dimensional context
    embedding comprising: a three-dimensional one-hot target index (from the LLM),
    the $x$-$y$ positions of all three targets ($6$D), and the robot's
    current position and yaw ($3$D).
  \item \textbf{VLA-based \modC:} A SmolVLA backbone~\citep{shukor2025smolvla}
    maps the image and language input to a learned $16$-dimensional context
    embedding $z_t \in [-1,1]^{16}$ via an MLP adapter applied to the
    representation immediately before the VLA action head.
\end{enumerate}

\noindent\textbf{Reward and termination.}
The agent receives $+1.0$ for reaching the correct goal, $-0.2$ for any
collision, and a step penalty of $-0.01$.
Episodes terminate upon goal completion, a collision, or after $T{=}500$ steps.

\subsection{Outdoor Navigation}
\label{app:env_outdoor}

\noindent\textbf{Setup.}
Outdoor Navigation is a target-reaching task evaluated on a Unitree GO2 quadruped~\citep{rudin2022learning} in a $16\times16$\,m simulated outdoor workspace.

\noindent\textbf{Observations.}
The agent receives a bird's-eye-view RGB image $I_t \in \R^{384\times384\times3}$
encoding the scene, a one-hot target index $g \in \{0,1,2\}$ identifying the
destination, and the safety observation:
\[
  s_t = [s_x,\, s_y,\, s_\psi] \in [0,1]^3,
\]
where $s_x$ and $s_y$ are the normalized planar position and
$s_\psi = (\psi + 180^\circ)/360^\circ$ is the normalized yaw.

\noindent\textbf{Action space.}
$\mathcal{A} = \{\textsc{Forward}(0),\,\textsc{Backward}(1),\,
\textsc{TurnLeft}(2),\,\textsc{TurnRight}(3)\}$.
Each discrete action maps to a velocity command $a_t=(v_x,\omega)$, where $v_x$
denotes forward linear velocity and $\omega$ yaw rate, clipped to $[-1,1]^2$ and
executed for $0.5$\,s:
\[
\begin{aligned}
  \textsc{Forward}(0) &\mapsto (1,0),\\
  \textsc{Backward}(1) &\mapsto (-1,0),\\
  \textsc{TurnLeft}(2) &\mapsto (0,1),\\
  \textsc{TurnRight}(3) &\mapsto (0,-1).
\end{aligned}
\]

\noindent\textbf{Dynamics.}
Each high-level action is a macro-action held for
$\Delta t_{\mathrm{high}}=0.5$\,s, implemented as $N{=}25$ low-level control
updates with $\Delta t_{\mathrm{low}}=0.02$\,s
(so $\Delta t_{\mathrm{high}}=N\,\Delta t_{\mathrm{low}}$).
The macro-action is realized by a pretrained low-level locomotion policy,
trained following~\cite{rudin2022learning}, which maps the clipped command
$a_t=(v_x,\omega)$ to joint-angle targets.
We assume the low-level controller reliably completes the corresponding movement. The updated position and orientation reflect the intended motion, abstracting away motor-level details such as joint angles.

\noindent\textbf{Reward and termination.}
The agent receives $+1.0$ for reaching the correct target, $+0.1 \cdot \Delta d_t$ to encourage progress toward the goal, where $\Delta d_t \doteq d_{t-1} - d_t$ and $d_t$ is the distance to the target at time $t$, $-1.0$ for colliding with a wall, and a step penalty of $-0.01$ to encourage efficiency.
Episodes terminate on success or boundary contact, or truncate after
$T{=}40$ high-level steps ($20$\,s).

\section{Language Descriptions}
\label{app:language_prompts}

This section lists the language prompts used in the Playground environment. The prompts describe the same three target objects using different phrasings, allowing us to evaluate whether the policy can ground semantically similar instructions to the correct goal.

\begin{promptbox}{}
\textbf{Target 0: Red circle}
\begin{enumerate}[leftmargin=20pt, itemsep=2pt, topsep=3pt]
    \item Reach the red circle with a black number zero (0) inside it.
    \item Move to the red circle labeled 0.
    \item Navigate to the red target marked with zero.
    \item Go toward the red goal with the digit 0.
    \item Your task is to touch the red circle labeled 0.
    \item Approach the red dot containing a black 0.
    \item Find and reach the red zero-marked circle.
    \item Drive the agent to the red goal labeled zero.
    \item Head to the red circle with a number zero inside.
    \item Move the agent to the red destination marked 0.
\end{enumerate}

\medskip

\textbf{Target 1: Green circle}
\begin{enumerate}[leftmargin=20pt, itemsep=2pt, topsep=3pt]
    \item Reach the green circle with a black number one (1) inside it.
    \item Move to the green circle labeled 1.
    \item Navigate to the green target marked with one.
    \item Go toward the green goal with the digit 1.
    \item Your task is to touch the green circle labeled 1.
    \item Approach the green dot containing a black 1.
    \item Find and reach the green one-marked circle.
    \item Drive the agent to the green goal labeled one.
    \item Head to the green circle with a number one inside.
    \item Move the agent to the green destination marked 1.
\end{enumerate}

\medskip

\textbf{Target 2: Blue circle}
\begin{enumerate}[leftmargin=20pt, itemsep=2pt, topsep=3pt]
    \item Reach the blue circle with a black number two (2) inside it.
    \item Move to the blue circle labeled 2.
    \item Navigate to the blue target marked with two.
    \item Go toward the blue goal with the digit 2.
    \item Your task is to touch the blue circle labeled 2.
    \item Approach the blue dot containing a black 2.
    \item Find and reach the blue two-marked circle.
    \item Drive the agent to the blue goal labeled two.
    \item Head to the blue circle with a number two inside.
    \item Move the agent to the blue destination marked 2.
\end{enumerate}
\end{promptbox}

\section{Safety Specifications}
\label{app:safety_specs}

Safety properties are expressed as input-domain-conditioned output constraints
on \modS: for a specified region $\mathcal{R} \subset \mathcal{S}$ of the
safety-sensor space, certain actions are declared unsafe and must not be selected
by \modS for \emph{any} context embedding $z_t \in \mathcal{Z}$.
The formal statement is $s_t \in \mathcal{R} \Rightarrow
\pi_S(\cdot \mid x_t)\notin\mathcal{Y}_\text{unsafe}(\mathcal{R})$.

\paragraph{Playground (4 properties).}
The safety variable is the normalized position $s_t=(s_x,s_y)\in[0,1]^2$.
Boundary constraints prevent actions that move the agent further toward a
boundary when it is already within $10\%$ of that boundary:
\[
\begin{aligned}
s_x \in [0.9,1] &\;\Rightarrow\; \neg \textsc{Right}, \\
s_x \in [0,0.1] &\;\Rightarrow\; \neg \textsc{Left}, \\
s_y \in [0.9,1] &\;\Rightarrow\; \neg \textsc{Up}, \\
s_y \in [0,0.1] &\;\Rightarrow\; \neg \textsc{Down}.
\end{aligned}
\]

\paragraph{Indoor Navigation (5 properties).}
The safety variable is the 11-ray LiDAR scan
$s_t \in [0,3]^{11}$ ordered from left to right over a $180^\circ$ forward
field of view (ray indices $1\text{--}11$; smaller values indicate nearby obstacles).
The five properties encode directional obstacle avoidance;
Table~\ref{tab:stretch_props} lists the full specification.

\begin{table}[ht]
\centering
\caption{Indoor Navigation safety properties: LiDAR-conditioned unsafe actions.
Ray ranges not listed are unconstrained (full range $[0,3]$).}
\label{tab:stretch_props}
\setlength{\tabcolsep}{4pt}
\small
\begin{tabular}{clc}
\toprule
Prop. & Constrained LiDAR rays (threshold $0.2$\,m) & Unsafe action(s) \\
\midrule
1 & $s_{1\text{--}3}\in[0,0.2]$ & \textsc{Turn-Left} (0), \textsc{Forward} (2) \\
2 & $s_{4\text{--}5}\in[0,0.2]$ & \textsc{Turn-Left} (0), \textsc{Forward} (2) \\
3 & $s_{5\text{--}7}\in[0,0.2]$ & \textsc{Forward} (2) \\
4 & $s_{7\text{--}8}\in[0,0.2]$ & \textsc{Turn-Right} (1), \textsc{Forward} (2) \\
5 & $s_{9\text{--}11}\in[0,0.2]$ & \textsc{Turn-Right} (1), \textsc{Forward} (2) \\
\bottomrule
\end{tabular}
\end{table}

The same five properties are applied to both the custom MDP and the VLA-based
\modC variants, as well as to the mapless POMDP variant used for sim-to-real
transfer.

\paragraph{Outdoor Navigation (10 properties).}
The safety variable is the normalized planar pose
$s_t=[s_x,s_y,s_\psi]\in[0,1]^3$.
Translational safety is heading-dependent: near each of the four workspace
boundaries, the translational action that moves the robot further outward
(determined by whether the robot is facing toward or away from the boundary)
is declared unsafe.
Turning actions are never restricted.
Table~\ref{tab:unitree_props} lists all ten properties.

\begin{table}[h]
\centering
\caption{Outdoor Navigation safety properties: boundary- and orientation-conditioned
unsafe translational actions. $s_\psi = (\psi+180^\circ)/360^\circ$; $s_x,s_y$
are normalized to $[0,1]$ over the $16$\,m workspace.}
\label{tab:unitree_props}
\setlength{\tabcolsep}{4pt}
\small
\begin{tabular}{cllc}
\toprule
Prop. & Position range & Heading range & Unsafe action \\
\midrule
1  & $s_x\in[0,0.125]$ & $s_\psi\in[0.875,1.0]$     & \textsc{Forward} (0) \\
2  & $s_x\in[0,0.125]$ & $s_\psi\in[0,0.125]$        & \textsc{Forward} (0) \\
3  & $s_x\in[0,0.125]$ & $s_\psi\in[0.375,0.625]$    & \textsc{Backward} (1) \\
4  & $s_x\in[0.875,1]$ & $s_\psi\in[0.375,0.625]$    & \textsc{Forward} (0) \\
5  & $s_x\in[0.875,1]$ & $s_\psi\in[0.875,1.0]$      & \textsc{Backward} (1) \\
6  & $s_x\in[0.875,1]$ & $s_\psi\in[0,0.125]$        & \textsc{Backward} (1) \\
7  & $s_y\in[0,0.125]$ & $s_\psi\in[0.125,0.375]$    & \textsc{Forward} (0) \\
8  & $s_y\in[0,0.125]$ & $s_\psi\in[0.625,0.875]$    & \textsc{Backward} (1) \\
9  & $s_y\in[0.875,1]$ & $s_\psi\in[0.625,0.875]$    & \textsc{Forward} (0) \\
10 & $s_y\in[0.875,1]$ & $s_\psi\in[0.125,0.375]$    & \textsc{Backward} (1) \\
\bottomrule
\end{tabular}
\end{table}

The boundary threshold of $0.125$ (normalized) corresponds to $2$\,m in the
$16$\,m workspace, providing sufficient margin for the robot to decelerate and
reorient before reaching the boundary.

\section{Architecture and Training Procedure}
\label{app:arch_training}

This section provides implementation details for all \modC instantiations and
training procedures used in the experiments.
In all cases, \modS is a two-layer MLP with $32$ hidden units per layer and a
$\tanh$ output activation bounding $z_t$ to $[-1,1]^{d_c}$.
The combined input to \modS is $x_t = [s_t;\, z_t]$, with $s_t$ the
safety sensor vector and $z_t$ the bounded context embedding from \modC.

\paragraph{Playground.}
\modC consists of a BERT text encoder~\citep{devlin2019bert} and a ViT image
encoder~\citep{dosovitskiy2020image}.
Both foundation models are fine-tuned jointly with \modS using PPO, with LoRA
applied to all attention layers.
The resulting context embedding is a free-form bounded vector
$z_t \in [-1,1]^{64}$.

\paragraph{Indoor Navigation — Custom \modC.}
The LLM and ViT components are trained separately by supervised fine-tuning
before \modS is trained with PPO.
The LLM is fine-tuned (with LoRA) on instruction--target-index pairs to predict
a one-hot target identifier from the language prompt.
The ViT is fine-tuned on image--label pairs generated by the simulator, where
each label contains the ground-truth positions of all three target regions, the
robot's 2D position, and its yaw angle.
The structured context embedding is formed by concatenating these outputs:
one-hot target index ($3$D), target positions ($6$D: three $(x,y)$ pairs), robot
position ($2$D), and robot yaw ($1$D), yielding $z_t \in \R^{12}$.
The $\tanh$ activation bounds $z_t$ component-wise to $[-1,1]^{12}$.
After supervised pre-training, \modS is trained with PPO on the fixed structured
embeddings.

\paragraph{Indoor Navigation — VLA-based \modC.}
\modC uses SmolVLA~\citep{shukor2025smolvla}, specifically the
\texttt{SmolVLM2-256M-Video-Instruct} backbone (${\approx}256$M parameters).
The feature representation immediately before the VLA action head is projected
through a two-layer MLP adapter to a $16$-dimensional bounded embedding
$z_t \in [-1,1]^{16}$.
The backbone is fine-tuned with LoRA applied to all language-model attention
layers (rank $r{=}8$, scaling factor $\alpha{=}16$, dropout $0.05$), while
the image processing layers are kept frozen.
The adapter and LoRA parameters, together with \modS, are trained end-to-end
with PPO.
This configuration evaluates whether the \fearl context interface can accommodate
a standard off-the-shelf VLA backbone.
pipeline.

\paragraph{Outdoor Navigation}
For Outdoor Navigation, we use a two-stage procedure: first, we train a compact MLP teacher with PPO using low-dimensional inputs, including the robot position, yaw, and target location; then, we distill this teacher into the final high-capacity FEARL policy, consisting of a ViT-based Controller and Module~S.
We train the student with DAgger~\cite{ross2011reduction}: the student collects trajectories under its own policy, and the teacher is queried at each visited state to provide the expert action. This mitigates covariate shift compared to standard behavior cloning and improves robustness during deployment.

\section{Training Hyperparameters}
\label{app:hyperparams}

Table~\ref{tab:hyperparams} summarizes the key architectural and training
hyperparameters across all evaluated configurations.

\begin{table}[ht]
\centering
\caption{Training configuration across environments. ``SFT'' denotes supervised.}
\label{tab:hyperparams}
\small
\setlength{\tabcolsep}{4.5pt}
\begin{tabular}{lcccccc}
\toprule
Configuration & \modS layers & \modS units & $d_s$ & $d_c$ & FM tuning & Episode $T$ \\
\midrule
Playground                 & 2 & 32 & 2  & 64  & LoRA         & 500 \\
Indoor Nav (Custom)        & 2 & 32 & 11 & 12  & SFT (+LoRA)  & 500 \\
Indoor Nav (VLA)           & 2 & 32 & 11 & 16  & LoRA (PPO)   & 500 \\
Outdoor Nav           & 2 & 32 & 2 & 128  & Dagger   & 40 \\
\bottomrule
\end{tabular}
\end{table}

For all PPO runs we use an episode horizon of $T$ steps (Table~\ref{tab:hyperparams}),
GAE with $\lambda{=}0.95$, discount $\gamma{=}0.99$, and a clipping ratio of
$\varepsilon{=}0.2$.
LoRA rank is $r{=}8$ with scaling $\alpha{=}16$ and dropout $0.05$ for all
foundation-model fine-tuning.
The \eprove verifier is configured with a point-cloud size of $3500$, a
maximum depth of $20$, targeting $\rho{=}0.5\%$ false positives at confidence
$\delta{=}99.99\%$ for all simulated environments.

\section{Sim-to-Real Implementation Details}
\label{app:sim2real}

This section describes the hardware setup and LiDAR preprocessing pipeline
used to transfer the Indoor Navigation (Mapless) policy from simulation to the
physical Hello Robot Stretch~2.

\subsection{Hardware Setup}

The robot is deployed in a structured indoor space of approximately
$2.5 \times 1.8$\,m, with three goal regions physically marked on the floor.
\begin{itemize}
  \item \textbf{Obstacle sensing.} The onboard 2D LiDAR provides a planar
    scan of $1080$ range measurements at $360^\circ/1080 \approx 0.333^\circ$
    angular resolution, at a range of up to $12$\,m.
  \item \textbf{Localization.} 
    An HTC Vive Tracker~2.0 mounted on the robot, together with two infrared base stations, provides 6-DoF pose estimates at approximately $100$\,Hz. We use only the tracker's planar position and yaw angle for policy input. These values are clipped to the arena bounds and normalized to $[0,1]$ using the arena dimensions, matching the preprocessing used during simulation training.
\end{itemize}

\subsection{LiDAR Preprocessing}

The FEARL safety module and shield expect an $11$-ray, $180^\circ$ front-facing
LiDAR vector in $[0,3]^{11}$, identical to the simulation representation.
We map the raw $1080$-ray full-circle scan to this compact representation through
the following steps.

\begin{enumerate}
  \item \textbf{Ray selection.} Eleven target ray angles are uniformly spaced
    over the $[-90^\circ, +90^\circ]$ frontal field of view.
    For each target angle we find the closest beam index in the raw scan using
    the scan's angle metadata.
  \item \textbf{Spatial filtering.} Rather than reading a single beam, each
    selected ray is computed by averaging valid returns within a small angular
    window (typically $\pm 2$ raw beams) around the target index.
    This reduces sensitivity to isolated invalid returns (e.g., LiDAR dropouts
    or specular reflections).
  \item \textbf{Temporal filtering.} The spatially filtered 11-ray vector is
    averaged over a short sliding window of the most recent valid scans
    (typically $3$--$5$ frames at $\approx10$\,Hz), further suppressing transient
    noise.
  \item \textbf{Missing-return handling.} If no valid reading is available for
    a given ray after spatial filtering, that ray is assigned the maximum range
    ($3.0$\,m).
  \item \textbf{Clipping.} All final range values are clipped to $[0, 3.0]$\,m,
    matching the training and verification domain.
\end{enumerate}

This preprocessing pipeline is the only hardware-specific modification applied
to the trained policy and shield.
Crucially, no retraining or re-verification is required: because the safety
module operates on the normalized LiDAR vector rather than raw sensor data,
the same verified shield can be reused on hardware as long as the preprocessed
observations remain within the bounds and normalization conventions used during
offline verification.

\subsection{Real-World Results}

Over $18$ deployment episodes, the shielded policy achieves a $61.1\%$ success
rate with zero collisions.
Disabling the shield while keeping all other components identical yields the
same $61.1\%$ success rate but a $27.8\%$ collision rate, confirming that the
safety mechanism is active and effective.
The shield intervenes in $7.8\%$ of steps, consistent with the $22.9\%$
uncertified volume identified offline (the uncertified fraction is an upper
bound on the shield's activation domain, not on the actual activation rate).
The lower success rate compared to simulation ($77.1\%$ certified in simulation
vs.\ $61.1\%$ real-world success) is attributed to sim-to-real distribution
shift in task-level inputs, primarily localization drift from the Vive tracker
and actuation noise.
Importantly, safety remains intact: the shield's LiDAR-based input is
insensitive to localization errors, as it depends only on the onboard obstacle
sensor rather than the full task observation.

\subsection{Language Generalization}

The real-world deployment uses language prompts not seen during simulation
training (e.g., paraphrased descriptions of the goal regions).
The LLM-only \modC successfully grounds these novel prompts to the correct
goal index in the majority of trials, demonstrating that the pretrained language
backbone retains useful generalization under the \fearl interface.
A representative failure case was observed when the Vive tracker produced an
incorrect robot pose, causing the task-level state estimate to diverge; despite
this, the robot consistently avoided obstacles, illustrating that the safety
shield is robust to failures in non-safety sensors.

\end{document}